\pdfoutput=1

\documentclass[11pt]{article}
\usepackage[table]{xcolor}

\usepackage{acl}

\usepackage{times}
\usepackage{latexsym}

\usepackage[T1]{fontenc}

\usepackage[utf8]{inputenc}

\usepackage{microtype}

\usepackage{url}
\usepackage{easyReview}

\usepackage{cleveref}
\usepackage{multirow}
\usepackage{tabularx}

\usepackage{graphicx}

\usepackage{subfig}
\usepackage{booktabs}

\usepackage{caption}
\usepackage{tikz}

\definecolor{cadetgrey}{rgb}{0.57, 0.64, 0.69}
\definecolor{greygrey}{rgb}{0.169,0.169,0.169}

\title{Probing Taxonomic and Thematic Embeddings \\ for Taxonomic Information}

\author{Filip Klubička \and John D. Kelleher \\
  ADAPT Centre, Technological University Dublin, Ireland \\
  \texttt{\{filip.klubicka,john.kelleher\}@adaptcentre.ie} \\}

\begin{document}
\maketitle
\begin{abstract}

Modelling taxonomic and thematic relatedness is important for building AI with comprehensive natural language understanding. The goal of this paper is to learn more about how taxonomic information is structurally encoded in embeddings. To do this, we design a new hypernym-hyponym probing task and perform a comparative probing study of taxonomic and thematic SGNS and GloVe embeddings. Our experiments indicate that both types of embeddings encode some taxonomic information, but the amount, as well as the geometric properties of the encodings, are independently related to both the encoder architecture, as well as the embedding training data. Specifically, we find that only taxonomic embeddings carry taxonomic information in their norm, which is determined by the underlying distribution in the data. 

\end{abstract}

\section{Introduction}

Research on probing \cite{ettinger-etal-2016-probing,shi-etal-2016-string,veldhoen2016diagnostic,adi2017fine} has gained significant momentum in the NLP community in recent years, helping researchers explore different aspects of text encodings. While its potential for application is broad, there are still many NLP tasks the framework has not been applied to. Specifically, it seems the majority of impactful probing work focuses on analysing syntactic properties encoded in language representations, yet the rich and complex field of semantics is comparably underrepresented \cite{belinkov-glass-2019-analysis}. One particular semantic problem that has not been explored at all in the context of probing is the distinction between the \textbf{taxonomic} and \textbf{thematic} dimensions of semantic relatedness \citep{magda2019thematic}: words or concepts which belong to a common taxonomic category share properties or functions, and such relationships are commonly reflected in knowledge-engineered resources such as ontologies or taxonomies. On the other hand, thematic relations exist by virtue of co-occurrence in a (linguistic) context where the relatedness is specifically formed between concepts performing complementary roles in a common event or theme. 

This distinction informs the theoretical basis of our work, as we wish to explore the tension between taxonomic and thematic representations by examining how their information is structurally encoded. Indeed, the vast majority of pretrained language models (PTLMs) are trained solely on natural language corpora, meaning they mainly encode thematic relations. 
Consequently, most probing work is applied to thematic embeddings, while taxonomic embeddings remain unexplored. We thus use the probing framework to study and compare taxonomic and thematic meaning representations. 

In addition, one aspect of embeddings that has not received much attention is the contribution of the vector norm to encoding linguistic information. We have recently highlighted this gap in the literature and developed an extension of the probing method called \textit{probing with noise} \citep{klubicka-kelleher-2022-probing}, which allows for relative intrinsic probe evaluations that are able to provide structural insights into embeddings and highlight the role of the vector norm in encoding linguistic information. We find taxonomic embeddings to be particularly interesting for probing the role of the norm, as we suspect that the hierarchical structure of a taxonomy is well suited to be encoded by the vector norm---given that the norm encodes the vector's magnitude, or distance from the space's origin, it is possible that the depth of a tree structure, such as a taxonomy, could be mapped to the vector's distance from the origin in some way\footnote{A hypothesis based on the finding that the squared L2 norm of BERT and ELMo can correspond to the depth of the word in a syntactic parse tree \cite{hewitt-manning-2019-structural}.}. Applying the \textit{probing with noise} method to taxonomic embeddings on a taxonomic probing task could shed some light on this relationship. In order to draw broader comparisons, we apply the same evaluation framework to taxonomic and thematic SGNS and GloVe embeddings.

\section{Related Work}

Hypernymy, understood as the capability to relate generic terms or classes to their specific instances, lies at the core of human cognition and plays a central role in reasoning and understanding natural language \citep{wellman1992cognitive}. Two words have a hypernymic relation if one of the words belongs to a taxonomic class that is more general than that of the other word. Hypernymy can be seen as an \textit{IS-A} relationship, and more practically, hypernymic relations determine lexical entailment \citep{geffet-dagan-2005-distributional} and form the \textit{IS-A} backbone of almost every ontology, semantic network and taxonomy \citep{yu2015learning}. Given this, it is not surprising that modelling and identifying hypernymic relations has been pursued in NLP for over two decades \citep{shwartz-etal-2016-improving}.

While research on hypernym detection has been plentiful, work applying any probing framework to identify taxonomic information in embeddings is scarce, and the existing work does nor probe for it directly, but rather infers taxonomic knowledge from examining higher-level tasks. For example, \citet{ettinger-2020-bert} identified taxonomic knowledge in BERT, but rather than using a probing classifier, BERT’s masked-LM component was used instead and its performance was examined on a range of cloze tasks. One of the relevant findings was that BERT can robustly retrieve noun hypernyms in this setting, demonstrating that BERT is strong at associating nouns with their hypernyms. \citet{ravichander2020systematicity} build on Ettinger's work and investigate whether probing studies shed light on BERT's systematic knowledge, and as a case study examine hypernymy information. They devise additional cloze tasks to test for prediction consistency and demonstrate that BERT often fails to consistently make the same prediction in slightly different contexts, concluding that its ability to correctly retrieve hypernyms is not a reflection of larger systematic knowledge, but possibly an indicator of lexical memorisation \citep{levy-etal-2015-supervised,santus-etal-2016-nine,shwartz-etal-2017-hypernyms}.

Aside from this recent focus on BERT, little work has been done in the space of probing embeddings for hypernym information. However, work on modelling hypernymy has a long history that stretches back before large PTLMs and includes pattern-based approaches \citep{hearst1992automatic,navigli-velardi-2010-learning,lenci-benotto-2012-identifying,boella-di-caro-2013-extracting,flati-etal-2014-two,santus-etal-2014-chasing,flati2016multiwibi,gupta-etal-2016-revisiting,pavlick-pasca-2017-identifying} that are based on the notion of distributional generality \citep{weeds-etal-2004-characterising, clarke-2009-context}, as well as distributional approaches \citep{Turney2010,baroni-etal-2012-entailment,rei-briscoe-2013-parser,santus-etal-2014-chasing,fu-etal-2014-learning,espinosa-anke-etal-2016-supervised,ivan2017well,nguyen-etal-2017-hierarchical,pinter-eisenstein-2018-predicting,bernier-colborne-barriere-2018-crim,nickel2018learning,roller-etal-2018-hearst,maldonado-klubicka-2018-adapt,cho2020leveraging,mansar2021finsim}. We highlight the work of \citet{weeds-etal-2014-learning}, who demonstrated that it is possible to predict a specific semantic relation between two words given their distributional vectors. Their work is especially relevant to ours as it shows that the nature of the relationship one is trying to establish between words informs the operation one should perform on their associated vectors, e.g. summing the vectors works well for a co-hyponym task. We consider this in \S\ref{s:dataset}.

In terms of evaluation benchmarks for modeling hypernymy, in most cases their design reduces them to binary classification \citep{baroni-lenci-2011-blessed,snow2005learning,boleda-etal-2017-instances,vyas-carpuat-2017-detecting}, where a system has to decide whether or not a hypernymic relation holds between a given candidate pair of terms. Criticisms to this experimental setting point out that supervised systems tend to benefit from the inherent modeling of the datasets in the task, leading to lexical memorization phenomena. Some attempts to alleviate this issue involve including a graded scale for evaluating the degree of hypernymy on a given pair \citep{vulic-etal-2017-hyperlex}, or reframing the task design as Hypernym Discovery \cite{espinosa-anke-etal-2016-supervised}. The latter addresses one of the main drawbacks of the binary evaluation criterion and resulted in the construction of a hypernym discovery benchmark covering multiple languages and knowledge domains \cite{camacho-collados-etal-2018-semeval}.

\section{Probing Dataset Construction}
\label{s:dataset}

\citet{conneau-etal-2018-cram} state that a probing task needs to ask a simple, non-ambiguous question, in order to minimise interpretability problems and confounding factors. While we acknowledge the hypernym discovery framing as an important benchmark, and the cloze tasks used by \citet{ettinger-2020-bert} as an enlightening probing scenario, we suspect neither is suitable for our probing experiments, for which we require a simpler task that more directly teases out the hypernym-hyponym relationship. 
We thus opt to construct a new taxonomic probing task: predicting which word in a pair is the hypernym, and which is the hyponym. 
This dataset is directly derived from WordNet \citep{Fellbaum1998} and contains all its hypernym-hyponym pairs. Thus each word pair shares only an immediate hypernym-hyponym relationship between the candidate words: a word in a pair can \textit{only} be a hyponym or hypernym of the other. 

However, in our experiments we wish to probe both taxonomic and thematic encoders. Given that we are mostly using pretrained thematic and taxonomic embeddings (see \S\ref{s:models}), their vocabulary coverage 
might vary dramatically. We wish to mitigate confounders by comparing like for like as much as possible, so to retain a higher integrity of interpretation when comparing models, we prune the dataset to only use the intersection of vocabularies of all the used models---we only include word pairs that have a representation for both candidate words in all the embedding models. 

Note here that one of the goals of our work is to use the \textit{probing with noise} method to learn about embeddings and the way they encode different types of information in vector space. We assert that a prediction of the relationship between a pair of words cannot be fairly done without the classifier having access to representations for both words in the pair. Yet, our probe is a classifier which can only take a single vector as input. Informed by the work of \citet{weeds-etal-2014-learning} we considered options such as averaging or summing the individual word vectors, but found that these were not suitable for our framing as they muddled the notion that the classifier is receiving two separate words as input. We instead concatenate the word vectors in question and pass a single concatenated vector to the classifier (similar to approaches used by \citet{adi2017fine}). 
This approach allows us to formulate the task as a positional classification task: given a pair of words, is the first one the hypernym or the hyponym of the other? We can then assign each instance in the corpus a binary label---0 or 1---representing the class of the first word in the pair. The probe can then predict if the left half of the vector is the hyponym (0) of the right half, or whether it is its hypernym (1).

Finally, the inherent tree structure of WordNet means that a smaller number of words will be hypernyms, while a larger number will be hyponyms. We want to avoid the probe memorising the subset of words more likely to be hypernyms, but rather to learn from information encoded in the (differences between) vectors themselves. In an attempt to achieve this, we balance out the ratio of class labels by duplicating the dataset and swapping the hypernym-hyponym positions and labels. Before duplicating, we also define a hold-out test set of 25,000 instances, so as to exclude the possibility of the same word pair appearing in both the train and test split---thus, the probe will be evaluated only on unseen instances. This duplication resulted in a final dataset of 493,494 instances, of which 50,000 comprise the test set and 443,494 comprise the training set. Here are some example instances: 

\begin{table}[h]
\scalebox{0.99}{
\centering
\begin{tabular}{l}

\textit{\textbf{0}, north, direction} \\ 
\textit{\textbf{1}, direction, north} \\
\textit{\textbf{0}, hurt, upset} \\
\textit{\textbf{1}, upset, hurt} \\

\end{tabular}}
\label{tab:noiseprobing}
\end{table}

\section{Experimental Setup}
\label{s:models}

\subsection{Chosen Embeddings}

In our experiments we probe taxonomic and thematic SGNS embeddings, and make an analogous comparison with taxonomic and thematic GloVe embeddings. Usually pretrained taxonomic embeddings are not as easy to come by as thematic ones, but fortunately we were able to include a set of freely available taxonomic embeddings that are based on a random walk algorithm over the WordNet taxonomy, inspired by the work of \cite{goikoetxea2015}. In short, the approach is to generate a pseudo-corpus by crawling the WordNet structure and outputting the lexical items in the nodes visited, and then running the word embedding training on the generated pseudo-corpus. Naturally, the shape of the underlying knowledge graph affects the properties of the generated pseudo-corpus, while the types of connections that are traversed will affect the kinds of relations that are encoded in this resource. A Python implementation has been made freely available\footnote{\url{https://github.com/GreenParachute/wordnet-randomwalk-python}} and the embeddings have been shown to encode taxonomic information \citep{klubicka2019synthetic}. Ultimately we chose these embeddings as they allow us to be methodologically consistent by creating taxonomic embeddings that employ the same encoder architectures used to obtain thematic embeddings.

\paragraph{word2vec (SGNS)} For \textit{taxonomic SGNS} representations\footnote{\url{https://arrow.dit.ie/datas/12/}} we opt for embeddings trained on the pseudo-corpus that yielded the highest Spearman correlation score on the wn-paths benchmark (introduced by  \citet{klubicka2020walk}), i.e. the corpus with 2 million sentences, with the walk going both ways and with a 2-word minimum sentence length. The lack of a directionality constraint provides higher vocabulary coverage and a smaller proportion of rare words, while the 2-word minimum sentence length limit ensures that we only have representations for words that are part of WordNet's taxonomic graph and have at least one hypernym-hyponym relationship, which makes them suitable for this task. For the \textit{thematic SGNS} embeddings we use a pretrained model, and opt for the gensim\footnote{\url{https://radimrehurek.com/gensim/}} word2vec implementation which was trained on a part of the Google News dataset (about 100 billion tokens) and contains 300-dimensional vectors for 3 million words and phrases\footnote{word2vec-google-news-300}. 

\paragraph{GloVe} To train \textit{taxonomic GloVe} embeddings, we use a popular Python implementation of the GloVe algorithm\footnote{\url{https://github.com/maciejkula/glove-python}}\textsuperscript{,}\footnote{We used the following training parameters: window=10, no\_components=300, learning\_rate=0.05, epochs=30, no\_threads=2. Any other parameters are left as default.} and, importantly, train it on the same 2m-both-2w/s pseudo-corpus as the above taxonomic SGNS was trained on\footnote{\url{https://arrow.dit.ie/datas/9/}}. For the \textit{thematic GloVe} embeddings we use the original Stanford pretrained GloVe embeddings\footnote{\url{https://nlp.stanford.edu/projects/glove/}}, opting for the larger common crawl model, which was trained on 840 billion tokens and contains 300-dimensional embeddings for a total of 2.2 million words. 

Note that when we concatenate the two word embeddings required for an instance in the train or test set, they become a 600-dimensional vector which is then passed on as input to the probe.

\subsection{Probing with Noise}

The method is described in detail in \citet{klubicka-kelleher-2022-probing}\footnote{Code available here: \url{https://github.com/GreenParachute/probing-with-noise}}: in essence it applies targeted noise functions to embeddings that have an ablational effect and remove information encoded either in the norm or dimensions of a vector.

We remove information from the norm (abl.N) by sampling random norm values and scaling the vector dimensions to the new norm. Specifically, we sample the L2 norms uniformly from a range between the minimum and maximum L2 norm values of the respective embeddings in our dataset\footnote{Thematic SGNS: [0.6854, 9.3121]\\
Taxonomic SGNS: [2.1666, 7.6483]\\
Thematic GloVe: [3.1519, 13.1196]\\
Taxonomic GloVe: [0.0167, 6.3104]}.

To ablate information encoded in the dimensions (abl.D), we randomly sample dimension values and then scale them to match the original norm of the vector. Specifically, we sample the random dimension values uniformly from a range between the minimum and maximum dimension values of the respective embeddings in our dataset\footnote{Thematic SGNS: [-1.5547, 1.7109]\\
Taxonomic SGNS: [-1.8811, 1.7843]\\
Thematic GloVe: [-4.2095, 4.0692]\\
Taxonomic GloVe: [-1.3875, 1.3931]}. We expect this to fully remove all interpretable information encoded in the dimension values, making the norm the only information container available to the probe. 

Applying both noise functions to the same vector (abl.D+N) should remove any information encoded in it, meaning the probe has no signal to learn from, a scenario equal to training on random vectors. 

Even when no information is encoded in an embedding, the train set may contain class imbalance, and the probe can learn the distribution of classes. To account for this, as well as the possibility of a powerful probe detecting an empty signal \citep{zhang-bowman-2018-language}, we need to establish informative random baselines against which we can compare the probe's performance. We employ two such baselines: (a) we assert a random prediction (\textit{rand.pred}) onto the test set, negating any information that a classifier could have learned, class distributions included; and (b) we train the probe on randomly generated vectors (\textit{rand.vec}), establishing a baseline with access only to class distributions. 

Importantly, while we use randomised baselines as a sense check, we use the vanilla SGNS and GloVe word embeddings in their respective evaluations as \textit{vanilla baselines} against which all of the introduced noise models are compared. Here, the probe has access to both dimension and norm information, as well as class distributions from the training set. However, given the lack of probing taxonomic embeddings in the literature, it is equally important to establish the vanilla baseline's performance against the random baselines: we need to confirm that the relevant information is indeed encoded somewhere in the embeddings.

Finally, to address the degrees of randomness in the method, we train and evaluate each model 50 times and report the average score of all the runs, essentially bootstrapping over the random seeds \citep{wendlandt-etal-2018-factors}. Additionally, we calculate a confidence interval (CI) to make sure that the reported averages were not obtained by chance, and report it alongside the results.

\subsection{Probing Classifier and Evaluation Metric}

The embeddings are used as input to a Multi-Layered Perceptron (MLP) classifier, which predicts their class labels. We used the scikit-learn MLP implementation \citep{scikit-learn} using the default parameters\footnote{activation='relu', solver='adam', max\_iter=200, hidden\_layer\_sizes=100, learning\_rate\_init=0.001, batch\_size=min(200,n\_samples), early\_stopping=False, weight init. $
W \sim \mathcal{N}\left(0, \sqrt{6/(fan_{in}+fan_{out})}\right)$ (scikit relu default). See:  \url{https://scikit-learn.org/stable/modules/generated/sklearn.neural_network.MLPClassifier.html}}. The choice of evaluation metric used to evaluate the probes is not trivial, as we want to make sure that it reliably reflects a signal captured in the embeddings, especially in an imbalanced dataset where the probe could learn the label distributions, rather than detect a true signal related to the probed phenomenon. Following our original approach \citep{klubicka-kelleher-2022-probing}, we use the AUC-ROC score\footnote{\url{https://scikit-learn.org/stable/modules/generated/sklearn.metrics.roc_auc_score.html}}, which is suited to reflecting the classifier's performance on both positive and negative classes.

\section{Experimental Results}

Experimental evaluation results for taxonomic and thematic embeddings on the hypernym-hyponym probing task are presented in Tables \ref{t:results-sgns-tax} and \ref{t:results-glove-tax}. Note that all cells shaded light grey belong to the same distribution as random baselines on a given task, as there is no statistically significant difference between the different scores; cells shaded dark grey belong to the same distribution as the vanilla baseline on a given task; and all cells that are not shaded contain a significantly different score than both the random and vanilla baselines, indicating that they belong to different distributions.

\paragraph{SGNS}

\begin{table}
\centering
\scalebox{0.95}{
\centering
\begin{tabular}{|l|c|c|c|c|}
\hline
\multicolumn{5}{|m{8.1em}|}{\textbf{SGNS}} \\
\hline
Model & \multicolumn{2}{c|}{\textbf{THEM}} & \multicolumn{2}{c|}{\textbf{TAX}} \\ 
 & auc & \textpm CI & auc & \textpm CI \\
\hline
rand. pred. & \cellcolor{cadetgrey!25} .5000 & \cellcolor{cadetgrey!25} .0009 & \cellcolor{cadetgrey!25} .4997 & \cellcolor{cadetgrey!25} .0009  \\
rand. vec. & \cellcolor{cadetgrey!25} .5001 & \cellcolor{cadetgrey!25} .0012 & \cellcolor{cadetgrey!25} .5001 & \cellcolor{cadetgrey!25} .0011 \\
\hline
vanilla & \cellcolor{greygrey!25} .9163 & \cellcolor{greygrey!25} .0004 & \cellcolor{greygrey!25} .9256 & \cellcolor{greygrey!25} .0003 \\
\hline
 abl. N & .9057 & .0004 & .9067 & .0005 \\
 abl. D & .5039 & .0008 & .5294 & .0010 \\
 abl. D+N & \cellcolor{cadetgrey!25} .4998 & \cellcolor{cadetgrey!25} .0010 & \cellcolor{cadetgrey!25} .5002 & \cellcolor{cadetgrey!25} .0009 \\
\hline
\end{tabular}}
\caption{Probing results on SGNS models and baselines. Reporting average AUC-ROC scores and confidence intervals (CI) of the average of all training runs.}
\label{t:results-sgns-tax}
\end{table}

Starting with \textit{thematic SGNS} (THEM), Table \ref{t:results-sgns-tax} shows that the random baselines perform comparably to each other, as would be expected, and their score indicates no ability to discriminate between the two classes. We can see that the vanilla representations significantly outperform the random baselines, indicating that at least some taxonomic information is encoded in the embeddings. 

The norm ablation scenario (abl.N) causes a statistically significant drop in performance when compared to the vanilla baseline. In principle, this indicates that some information has been lost. If instead of the norm, we ablate the dimensions (abl.D), we see a much more dramatic performance drop compared to vanilla, indicating that much more information has been removed. Unsurprisingly, the difference in the probe's performance when applying both noising functions (abl.D+N) compared to random baselines is not statistically significant, meaning there is no pertinent information left in these representations. Notably, once just the dimension container is ablated, its performance drops to extremely low levels and approaches random baseline performance, yet it does not quite reach it---as small as it is, the difference is statistically significant, indicating that not all information has been removed in this setting. While significant, given how minor this difference is, one might argue it does not convincingly indicate the norm's role in encoding taxonomic information. 

However, we observe a much crisper signal in the \textit{taxonomic SGNS} (TAX) results. The random baselines perform comparably, while the vanilla baseline significantly outperforms them, while also significantly outperforming the THEM vanilla baseline, confirming that the taxonomic embeddings encode more taxonomic information than thematic embeddings. The norm ablation scenario causes a statistically significant performance drop from vanilla, while ablating the dimension container yields a larger drop, but does not reach the random-like performance achieved when ablating both containers. Here the difference in scores between ablating just the dimensions and ablating both dimensions and norm is also significantly different from random, but notably also an order of magnitude larger than in the THEM example. This indicates that the taxonomic SGNS embeddings use the norm to encode taxonomic information more so than thematic ones.

\paragraph{GloVe}

\begin{table}
\centering
\scalebox{0.95}{
\centering
\begin{tabular}{|l|c|c|c|c|}
\hline
\multicolumn{5}{|m{8.1em}|}{\textbf{GloVe}} \\
\hline
Model & \multicolumn{2}{c|}{\textbf{THEM}} & \multicolumn{2}{c|}{\textbf{TAX}} \\ 
 & auc & \textpm CI & auc & \textpm CI \\
\hline
rand. pred. & \cellcolor{cadetgrey!25} .4999 & \cellcolor{cadetgrey!25} .0011 & \cellcolor{cadetgrey!25} .4998 & \cellcolor{cadetgrey!25} .0010  \\
rand. vec. & \cellcolor{cadetgrey!25} .5001 & \cellcolor{cadetgrey!25} .0010 & \cellcolor{cadetgrey!25} .5001 & \cellcolor{cadetgrey!25} .0008 \\
\hline
vanilla & \cellcolor{greygrey!25} .9327 & \cellcolor{greygrey!25} .0004 & \cellcolor{greygrey!25} .8824 & \cellcolor{greygrey!25} .0005 \\
\hline
 abl. N & .9110 & .0004 & .8435 & .0008 \\
 abl. D & \cellcolor{cadetgrey!25} .5002 & \cellcolor{cadetgrey!25} .0008 & .6621 & .0008 \\
 abl. D+N & \cellcolor{cadetgrey!25} .5000 & \cellcolor{cadetgrey!25} .0011 & \cellcolor{cadetgrey!25} .5006 & \cellcolor{cadetgrey!25} .0011 \\
\hline
\end{tabular}}
\caption{Probing results on GloVe models and baselines. Reporting average AUC-ROC scores and confidence intervals (CI) of the average of all training runs.} 
\label{t:results-glove-tax}
\end{table}

In Table \ref{t:results-glove-tax} we see that \textit{thematic GloVe} (THEM) vanilla performance dramatically outperforms the baselines, but the scores drop when the norm is ablated. After ablating the dimension information, there is a substantial drop in the probe's performance and it is immediately comparable to random baselines with no statistically significant difference. Furthermore, performance does not significantly change after also ablating the norm. 

Meanwhile, the \textit{taxonomic GloVe} embeddings tell a different story. Firstly, while vanilla embeddings outperform the random baselines, they perform much worse than THEM vanilla GloVe, indicating an inferior representation for the hypernym-hyponym prediction task, even though they were trained on WordNet random walk pseudo-corpora (we discuss this in \S\ref{s:tax_discussion}). Ablating the dimensions causes a significant drop in performance, but it is nowhere near the random performance reached when ablating both dimensions and norm. This is a really strong signal that indicates the norm encodes some hypernym-hyponym information. This echoes the findings on SGNS, showing that taxonomic embeddings tend to use the norm to encode taxonomic information more so than thematic ones.
 
\subsection{Dataset Validation Experiments: Dimension Deletions}
\label{s:hyper-hypo_deletions}

Our experimental design is based on the assumption that providing the probe with a concatenated vector of word embeddings would allow it to infer the asymmetric relationship between the words and use that signal to make predictions. While we have taken some steps to ensure this and mitigate lexical memorisation (see \S\ref{s:dataset}), there is still a concern that the models could have memorised other regularities encoded in the individual word representations and used that information to make predictions. For example, while many candidate words can indeed be both hyponyms or hypernyms, given the tree structure of the taxonomy and the distribution of edges, the frequencies at which a word takes on a hypernym or hyponym role are still skewed. It is thus more likely that any given word will be a hyponym than a hypernym, and it is possible that the embeddings implicitly encode the frequency at which a word takes on a hypernym role, versus a hyponym role.

To validate that the probe is actually learning a relationship between the candidate words, we run an additional batch of probing experiments to establish another set of baselines specific to this particular probing task. We examine the impact of two scenarios on the probe's performance: given the same labels, a) what if the probe's input was only one word vector, and b) what if the probe's input was only half of each word vector in the pair?

We denote this line of enquiry as \textit{deletion experiments}, given that in practice a) can be seen as deleting half of the concatenated vector, and b) as deleting one half each vector before concatenating. The crucial difference is that in a) the probe can only learn from one word vector without having any access to a representation of the other word, meaning it can only predict whether the candidate word is a hyponym or a hypernym by relying on the probability derived from its frequency. In b) the probe has a representation for both vectors, meaning it could leverage the relationship between them, but the individual vectors are truncated, meaning that half of the dimensions are gone for each word, making this inferior to the vanilla setting\footnote{This choice is motivated by a desire to make this setting comparable to a) in terms of dimensionality---had we simply compared it to vanilla, it would have the advantage of having access to twice as many dimensions.}.

We ran these experiments for taxonomic and thematic SGNS and GloVe embeddings and when performing deletions assessed the impact of both halves of the vectors. All dimension deletion results are included in Tables \ref{t:results-del-sgns-tax} and \ref{t:results-del-glove-tax}, where scenario a) is denoted as \textit{del.ct.1h/2h} (deleted 1st/2nd half of concatenated vector) and scenario b) is denoted as \textit{del.ea.1h/2h} (deleted 1st/2nd half of each vector). When comparing the deletions of the different halves, in cases where there is a statistically significant difference between their scores, the lower of the two scores is marked with an asterisk (*). 

\paragraph{SGNS}

\begin{table}
\centering
\scalebox{0.95}{
\centering
\begin{tabular}{|l|c|c|c|c|}
\hline
\multicolumn{5}{|m{8.1em}|}{\textbf{SGNS}} \\
\hline
Model & \multicolumn{2}{c|}{\textbf{THEM}} & \multicolumn{2}{c|}{\textbf{TAX}} \\ 
 & auc & \textpm CI & auc & \textpm CI \\
\hline
rand. pred. & \cellcolor{cadetgrey!25} .5000 & \cellcolor{cadetgrey!25} .0009 & \cellcolor{cadetgrey!25} .4997 & \cellcolor{cadetgrey!25} .0009  \\
rand. vec. & \cellcolor{cadetgrey!25} .5001 & \cellcolor{cadetgrey!25} .0012 & \cellcolor{cadetgrey!25} .5001 & \cellcolor{cadetgrey!25} .0011 \\
\hline
vanilla & \cellcolor{greygrey!25} .9163 & \cellcolor{greygrey!25} .0004 & \cellcolor{greygrey!25} .9256 & \cellcolor{greygrey!25} .0003 \\
\hline
del. ea. 1h & .8929 & .0004 & .8998* & .0005 \\
del. ea. 2h & .8927 & .0004 & .9039 & .0004 \\
\hline
del. ct. 1h & .8496 & .0004 & .8525 & .0004 \\
del. ct. 2h & .8495 & .0004 & .8523 & .0003 \\
\hline
\end{tabular}}
\caption{Probing results on SGNS deletions and baselines. Reporting average AUC-ROC scores and confidence intervals (CI) of the average of all training runs.} 
\label{t:results-del-sgns-tax}
\end{table}

Unsurprisingly, deleting half of the vector in either scenario causes a statistically significant drop in performance when compared to vanilla. We also observe a larger drop in both \textit{del.ct}. settings versus the \textit{del.ea}. settings, which confirms that predicting a word's relationship to an ``imaginary'' other word is the more difficult task.

However, strikingly, the performance is also significantly above random, which indicates that the probe likely did learn some frequency distributions from the graph. It is possible that this is a reflection of the imbalance inherent to WordNet, given the large number of leaf nodes in the taxonomic graph.

Even still, the significant difference in scores between the two settings demonstrates that having access to both words, even at the cost of half the information in each word's dimensions, is more informative than having a full representation of a single word, \textit{indicating that the probe is inferring the relevant relationship between them}.

\paragraph{GloVe}

\begin{table}
\centering
\scalebox{0.95}{
\centering
\begin{tabular}{|l|c|c|c|c|}
\hline
\multicolumn{5}{|m{8.1em}|}{\textbf{GloVe}} \\
\hline
Model & \multicolumn{2}{c|}{\textbf{THEM}} & \multicolumn{2}{c|}{\textbf{TAX}} \\ 
 & auc & \textpm CI & auc & \textpm CI \\
\hline
rand. pred. & \cellcolor{cadetgrey!25} .4999 & \cellcolor{cadetgrey!25} .0011 & \cellcolor{cadetgrey!25} .4998 & \cellcolor{cadetgrey!25} .0010  \\
rand. vec. & \cellcolor{cadetgrey!25} .5001 & \cellcolor{cadetgrey!25} .0010 & \cellcolor{cadetgrey!25} .5001 & \cellcolor{cadetgrey!25} .0008 \\
\hline
vanilla & \cellcolor{greygrey!25} .9327 & \cellcolor{greygrey!25} .0004 & \cellcolor{greygrey!25} .8824 & \cellcolor{greygrey!25} .0005 \\
\hline
del. ea. 1h & .9120* & .0003 & .8727 & .0005 \\
del. ea. 2h & .9179 & .0004 & .8730 & .0006 \\
\hline
del. ct. 1h & .8522 & .0004 & .8405 & .0004 \\
del. ct. 2h & .8522 & .0004 & .8406 & .0004 \\
\hline
\end{tabular}}
\caption{Probing results on GloVe deletions and baselines. Reporting average AUC-ROC scores and confidence intervals (CI) of the average of all training runs.} 
\label{t:results-del-glove-tax}
\end{table}

The GloVe deletion results echo the findings on SGNS in most settings. Deleting half of the vector in either scenario causes a significant performance drop, which is largely above random performance, and the drop is larger in the \textit{del.ct.} setting versus the \textit{del.ea.} setting. This provides further indication that, while there is an inherent imbalance in the underlying data, the probe is inferring the relevant relationship between the candidate words when given a concatenation of two word vectors. The probe benefits significantly from having access to a representation of both words, or even just two halves of each representation. Even when it is not explicitly told that it is actually getting two inputs, it is able to pick up on the fact that there is a difference between them which can be helpful in deciding on a label.

\section{Discussion}
\label{s:tax_discussion}

There are a number of points to take away from our experimental results. Firstly, we see that both vanilla thematic embeddings encode taxonomic information and the GloVe vanilla model significantly outperforms the SGNS vanilla model. This is at least partially due to the fact that the pretrained SGNS and GloVe thematic embeddings were trained on unrelated corpora, which differ in terms of size, topic and coverage: the corpus that GloVe was trained on is over 8 times larger than the one used to train the SGNS model, and belongs to a different, much more varied genre of text data. Thus, word representations derived from these resources are likely very different and it is possible that due to the broader scope and much larger size of the GloVe corpus, the GloVe representations reflect more taxonomic knowledge.

However, these encoders exhibit the opposite behaviour when trained on the same WordNet random walk pseudo-corpus: expectedly, vanilla taxonomic SGNS scores improve upon its thematic version, yet vanilla taxonomic GloVe scores significantly underperform compared to thematic. While we would expect it to mirror what was observed in SGNS, taxonomic GloVe is in fact our worst-performing vanilla model. Given the significant differences in model architectures, it is possible that this unexpected behaviour is due to an interaction between the architecture and training data\footnote{The interested reader might consult \citet[pages 121-123]{klubicka-2022-thesis} for some speculation as to what that interaction might be.}. While this may play a role, we suspect that the dominant factor is rather training corpus size. The WordNet pseudo-corpus used for training taxonomic embeddings was only about 9 million tokens in size (which is sufficient to encode taxonomic relations, as shown by \citet{maldonado2019size}), whereas SGNS and GloVe were trained on 100 and 840 billion tokens respectively. It is not surprising that GloVe trained on a small and relatively sparse pseudo-corpus underperforms compared to training on a large natural corpus. If anything, it is encouraging that SGNS trained on a 9-million-token pseudo-corpus outperforms one trained on a 100-billion-token natural corpus.

Another important finding from our experiments is the strong evidence that \textit{word embedding models can use the norm to encode taxonomic information, regardless of what is encoded in the vector dimensions}. We find the clearest example of this in taxonomic GloVe after ablating dimension information, where the score remains as high as $\approx$0.66, meaning that the difference of 0.16 points is solely due to information in the norm. This is a very large difference given our understanding of the underlying mechanics, where it is well known that dimensions contain most, if not all information relevant for a task (e.g. \citet{durrani-etal-2020-analyzing,durrani2022linguistic}), and this is much more than has been demonstrated on any of the sentence-level experiments in our previous work \citep{klubicka-kelleher-2022-probing}. Additionally, this is the only case where deleting half of each word vector yields a significantly higher score ($\approx$0.87) than ablating the norm ($\approx$0.84). This suggests that more information is lost when the norm is ablated than when half of the dimensions are removed. This is a strong indicator that in this case the \textit{norm encodes information that is not at all available in the dimensions}. Certainly, the majority of the information in an embedding is and will always be encoded in the dimensions, but it is striking how much of it is present in the norm in this case.

Generally, when it comes to dimension deletion experiments, it is expected that the performance would drop dramatically in comparison to vanilla embeddings. However, an important takeaway is that in all settings the drop is much smaller than might be expected, being quite close to vanilla performance and largely above random performance. This points to a redundancy within the dimensions themselves, seeing as either half of the vector seems to carry more than half the information required to model the task, indicating that not many dimensions are needed to encode specific linguistic features. This is consistent with the findings of \cite{durrani-etal-2020-analyzing}, who analysed individual neurons in PTLMs and found that small subsets of neurons are sufficient to predict certain linguistic tasks. Our deletion results certainly corroborate these findings, given how small the drop in the probe's performance is when half the vector is deleted.

\begin{figure}[!h]
\centering
  \includegraphics[width=1.01\linewidth]{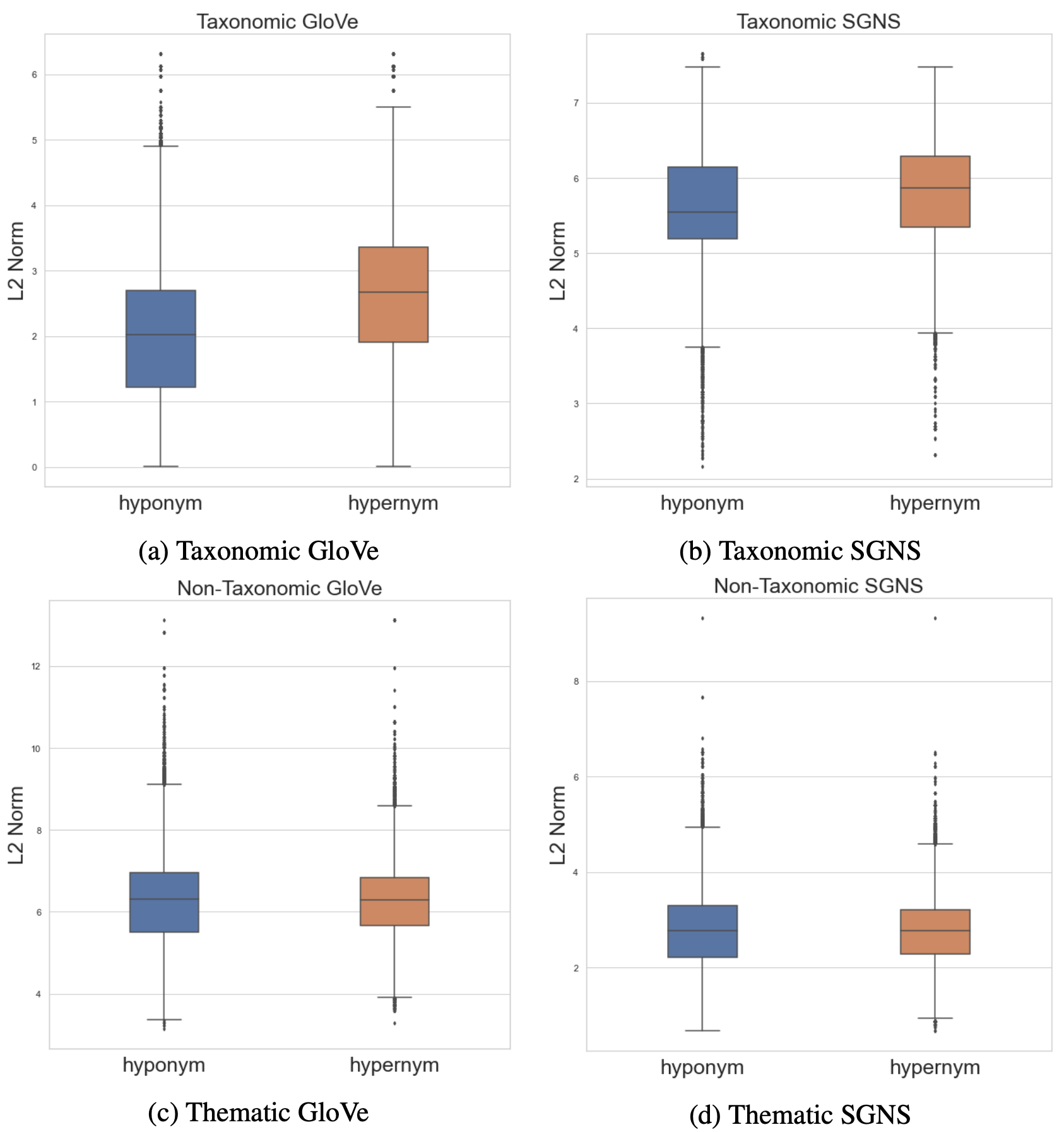}
\caption{Box plots depicting the median values of the L2 norm in the different sets of word vectors, separate for hyponyms and hypernyms. There is a marked difference observed between hyponym and hypernym norms in taxonomic GloVe and SGNS, but not in thematic.}
\label{f:box_norms}
\end{figure}

For additional insight into the norm, we examine the norm values. We calculate the norms of the individual hypernym and hyponym word vectors in our dataset and present the results in Figure \ref{f:box_norms}. The median norm value shows that the difference between hypernym and hyponym norms seems to be minor in both thematic embedding types (GloVe: 6.26 and 6.24; SGNS: 2.78 and 2.76), whereas the difference is an order of magnitude larger in both taxonomic representations (GloVe: 2.03 and 2.67; SGNS: 5.64 and 5.80). The difference is also quite large between taxonomic GloVe and SGNS, and it seems to be what is reflected in our experimental results, which show that GloVe stores the most hypernym-hyponym information in the norm.

The median norm measurements show that, on average, the norm of hypernyms is larger than the norm of hyponyms. This means that hypernyms, which are higher up in the tree, are positioned further away from the origin  of the vector space than hyponyms, which are positioned lower in the tree and are closer to the origin. Notably, this is only true in taxonomic embeddings, but not the thematic ones, indicating that in taxonomic embeddings \textit{there is a mapping between the taxonomic hierarchy and distance from the origin}.

Finally, in spite of the fact that taxonomic GloVe (TAX) is the worst-performing vanilla model, it is interesting that its norm also encodes the most taxonomic information. We base our interpretation of this result on the following: i) in many embeddings there is a high correlation between the norm and word frequency \citep{goldberg2017neural}, and ii) WordNet pseudo-corpora reflect hypernym-hyponym frequencies and co-occurrences. We suspect the principal signal that plays a role in the way taxonomic embeddings encode taxonomic knowledge is precisely these word co-occurrences, which GloVe is designed to capture. In turn, the norm can be seen as analogous to the hierarchical nature of taxonomic relationships and becomes the most accessible place to store this information. The thematic corpora reflect thematic co-occurrences and frequencies and hence GloVe (THEM) does not store taxonomic information in the norm, as such relations are not hierarchical in nature.

\section{Conclusion}

In this paper we applied the \textit{probing with noise} method to two different types of word representations---taxonomic and thematic---each generated by two different embedding algorithms---SGNS and GloVe---on a newly-designed taxonomic probing task. The overall findings are that (a) both taxonomic and thematic static embeddings encode taxonomic information, (b) that the norm of static embedding vectors carries some taxonomic information and (c) thus the vector norm is a separate information container at the word level. (d) While in some cases there can be redundancy between the information encoded in the norm and dimensions, at other times the norm can encode information that is not at all available in the dimensions, and (e) whether the norm is utilised at all is sometimes dependant on training data, not just the encoder architecture. 

We also show that in the case of SGNS, taxonomic embeddings outperform thematic ones on the task, demonstrating the usefulness of taxonomic pseudo-corpora in encoding taxonomic information. Indeed, this work serves to further emphasise the importance of the norm, showing that the taxonomic embeddings use the norm to supplement their encoding of taxonomic information. In other words, random walk corpora can improve taxonomic information in word representations, which is not always the case for natural language corpora. 

\section*{Acknowledgements}

This research was conducted with the financial support of Science Foundation Ireland under Grant Agreements No. 13/RC/2106 and 13/RC/2106\_P2 at the ADAPT SFI Research Centre at Technological University Dublin. ADAPT, the SFI Research Centre for AI-Driven Digital Content Technology, is funded by Science Foundation Ireland through the SFI Research Centres Programme, and is co-funded under the European Regional Development Fund.

\bibliography{anthology,custom,gwc2023-f} 
\bibliographystyle{acl_natbib}




\end{document}